\documentclass[]{llncs}
\usepackage{graphicx}
\usepackage{amsmath,amssymb} 
\usepackage{color}
\usepackage[width=122mm,left=12mm,paperwidth=146mm,height=193mm,top=12mm,paperheight=217mm]{geometry}
\usepackage{epsfig}
\usepackage{graphicx}
\usepackage[caption=false]{subfig}
\usepackage{multirow}

\usepackage{enumitem}

\begin{document}
\pagestyle{headings}
\mainmatter

\title{ Reasoning and Algorithm Selection Augmented Symbolic Segmentation} 



\author{Martin Lukac$^1$
\and
Kamila Abdiyeva$^1$
\institute{
	Nazarbayev University, Kazahkstan\\
	{\tt\small kabdiyeva@nu.edu.kz}}
\and
	Michitaka Kameyama$^2$
	\institute{
	Nazarbayev University, Kazahkstan\\
	{\tt\small martin.lukac@nu.edu.kz}\and Ishinomaki Senshu University, Japan\\
{\tt\small michikameyama@isenshu-u.ac.jp}
}}


\maketitle

\begin{abstract}
	In this paper we present an alternative method to symbolic segmentation: we approach symbolic segmentation as an algorithm selection problem. That is, let there be a set A of available algorithms for symbolic segmentation, a set of input features $F$, a set of image attribute $\mathbb{A}$ and a selection mechanism $S(F,\mathbb{A},A)$ that selects on a case by case basis the best algorithm. The semantic segmentation is then an optimization process that combines best component segments from multiple results into a single optimal result. The experiments compare three different algorithm selection mechanisms using three selected semantic segmentation algorithms. The results show that using the current state of art algorithms and relatively low accuracy of algorithm selection the accuracy of the semantic segmentation can be improved by 2\%. 
\end{abstract}

\section{Introduction}

The research field of computer vision contains currently several very hard open issues. One of the problems being investigated is the problem of the symbolic segmentation; in this task the algorithm must segment images into meaningful regions and then detect objects represented by these regions. Both segmentation and object recognition have been extensively studied using various approaches. For instance, for segmentation in various contexts several dedicated resources exist~\cite{martin:01,gelasca:08,everingham:10}. Similarly algorithms for various contexts have been developed such as for natural images~\cite{malik:99,shi:00,arbalez:06,maire:08}, for medical images~\cite{sharma:10,lathen:10,mharib:12,ali:14} or for biological images~\cite{ali:11,meijering:12}. The object recognition have received even more attention due to very high interest in computer vision from the industry. Some of the recent approaches to object recognition and detection include~\cite{lowe:99,heikkila:04,jung:06,felzenswalb:10,ciresan:12}. 

The combination of both segmentation and recognition is however more difficult and only recently larger amount of studies using Deep Learning methods significantly improved the state of art results~\cite{bharath:14,chen:14,long:15}. For instance semantic segmentation has been implemented as a combination of segmentation and recognition~\cite{carreira:12}, probabilistic models~\cite{tu:02,ladicky:10}, convolutional networks~\cite{bharath:14} or other approaches for either specific conditions~\cite{perera:06}, a unified framework~\cite{li:09} or interleaved recognition and segmentation~\cite{leibe:08}. Some of the main difficulties of semantic segmentation are:
\begin{enumerate}
	\item The segmentation by humans depends on recognition and higher level information~\cite{zavitz:14}
	\item The recognition is directly depending on features and regions from which the features are extracted.
	\item The context of the image strongly modulate segmentation and object recognition.
\end{enumerate}

As can be seen in computer science and other fields requiring algorithms it happens very often that several algorithms are implemented to solve similar or same problem in some varying contexts, environments or different types of inputs. The reason for such diversity and specificity is the fact that real-world problems are much more complex and dynamical than the current state of art software and hardware can handle. Consequently several approaches used the algorithm selection approach to improve the algorithms for various problems.

In this paper we propose the algorithm selection approach to the problem of symbolic segmentation. We base our work on previously proposed platform for algorithm selection in~\cite{lukac:13}. We show that using algorithm selection and high level reasoning about the results of algorithm processing allows to iteratively improve result of semantic segmentation. We analyze three different approaches for algorithm selection using either Bayesian Network (BN), Support Vector Machine (SVM) or a Neural Network (ANN). The main contributions of this paper are:
\begin{enumerate}
	\item Analysis of an iterative algorithm selection framework in the context of semantic segmentation
	\item Evaluation of three different machine learning approaches for semantic segmentation algorithms
	\item Demonstration of the fact that despite the low precision of the algorithm selector the resulting semantic segmentation is improved
\end{enumerate}

This paper is organized as follows. Section~\ref{sec:bck} introduces related and previous works and Section~\ref{sec:algosel} introduces the algorithm selection framework. Section~\ref{sec:exp} describes the experimentation and the results and Section~\ref{sec:con} concludes the paper and discusses future extensions.

\section{Previous Work and Background}
\label{sec:bck}
The general idea behind the algorithm selection is to select a unique algorithm for a particular set of properties, attributes and features  extracted from the data or obtained prior to processing. The algorithm selection was originally proposed by Rice~\cite{rice:76} for the problem of operating system scheduler selection. Since then the algorithm selection has been used in various problems but has never become a main stream of problem solving.

The reason for which algorithm selection is not a mainstream is dual: on one hand it is necessary to find distinctive features and on the other hand the problem studied should be difficult enough that extracting additional features from the input data is computationally advantageous. 

\begin{figure}[bht]
	\centering
	\includegraphics[width=0.7\linewidth]{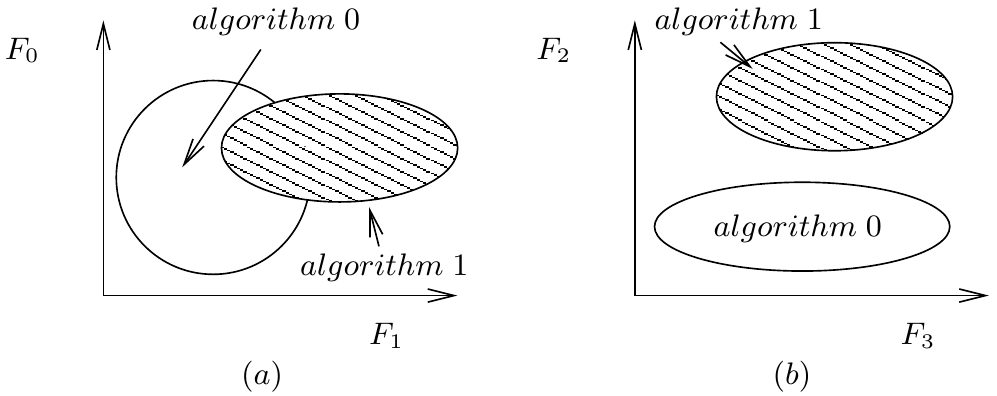}
	\caption{\label{fig:feats} Example illustrating (a) non-distinctive and (b) distinctive features}
\end{figure}

The concept of distinctive features is illustrated in Figure~\ref{fig:feats}. Figure~\ref{fig:feats}a shows that when features are not well identified the algorithm selection does not allow to uniquely determine the best algorithm because the features are non-distinctive for the available algorithms. Counter example using distinctive features is shown in Figure~\ref{fig:feats}b.

The ratio of computational effort that is required to extract additional features to the whole computation of the result can be estimated by comparing their respective computational time. In~\cite{lukac:12b} it was shown that for the task of image segmentation the algorithm selection is directly proportional to the size of the processed region of the image. If the region of segmentation is too small, the resulting segmentation of the tested algorithms results in very similar f-values and thus selecting fastest/computationally least expensive algorithm. For regions of larger size up to regions having the size of the input image, algorithm selection is both advantageous due to computational advantages as well as due to the increased quality of the result.

In computer vision and image processing the algorithm selection was previously on various levels of algorithmic processing. For instance, image segmentation of artificial~\cite{yong:03} or biological images~\cite{takemoto:09} was successfully implemented using algorithm selection approach. A set of features was found sufficient and allowed to clearly separate the area of performance of different algorithms. These two approaches however focused to separate the available algorithms only with respect to noise present in the image. Moreover, the algorithms used were single level line detectors such as Canny or the Prewitt. More complex algorithms for image segmentations were studied in~\cite{lukac:11d,lukac:12b}. Similarly to~\cite{yong:03,takemoto:09} a method using machine learning for algorithm selection for the segmentation of natural real-world images was developed. Other approaches have been studying the parameter selection or improving image processing algorithms using either machine learning or analytical methods but their approach is in general contained within a single algorithm~\cite{kolmogorov:07,peng:08,price:10}.

Methods and algorithms aimed at understanding of real world images have in general quite limited extend of their application. Currently there is a large amount of work combining segmentation and recognition and some of them are~\cite{ladicky:10,carreira:12}. In~\cite{leibe:08} an interleaved object recognition and segmentation is proposed in such manner that the recognition is used to seed the segmentation and obtain more precise detected objects contours. In~\cite{arbelaez:12} objects are detected by combining part detection and segmentation in order to obtain better shapes of objects. More general approaches such as~\cite{li:09} build a list of available objects and categories by learning them from data samples and reducing them to relevant information using some dictionary tool. However this approach does not scale to arbitrary size because the labels are not structured and ultimately require complete knowledge of the whole world.

In~\cite{hoiem:08} uses depth information to estimate whole image properties such as occlusions, background and foreground isolation and point of view estimation to determine type of objects in the image. All the modules of this approach are processed in parallel and integrated in a final single step. An airport apron analysis is performed in~\cite{ferryman:05} where the authors use motion tracking and understanding inspired by cognitive vision techniques. Finally, the image understanding can also be approached from a more holistic approach such as for instance in~\cite{oliva:01} where the intent is only to estimate the nature of the image and distinguish between mostly natural or artificial content.

\section{Algorithm Selection for Symbolic Segmentation}
\label{sec:algosel}
\begin{figure}[tbh]
\begin{center}
\fbox{
   \includegraphics[width=0.7\linewidth]{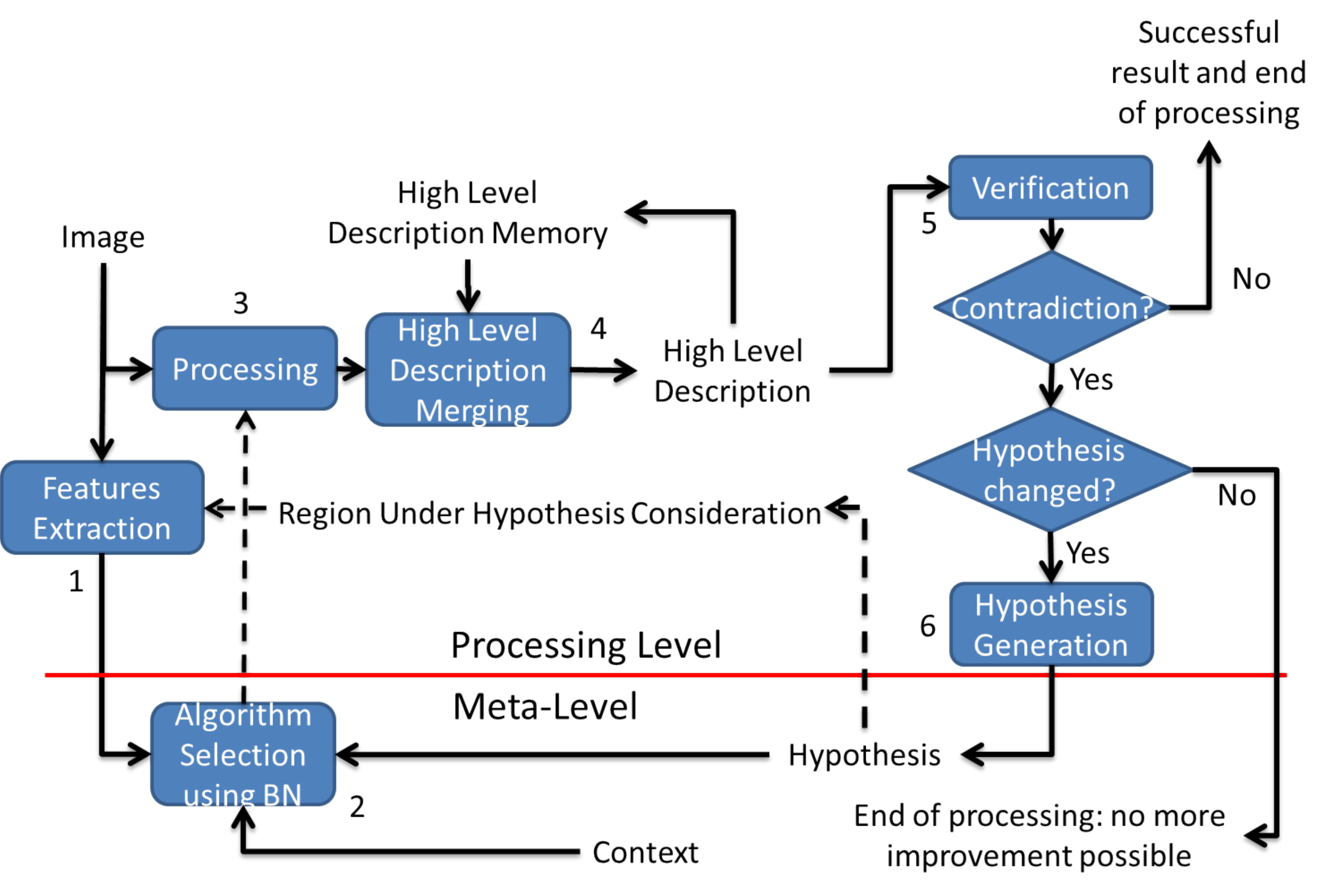}
}
\end{center}
   \caption{\label{fig:system}Algorithm Selection Platform}
\end{figure}

The framework used in these experiments was originally introduced in~\cite{lukac:13}. The schematic representation is shown in Figure~\ref{fig:system}. The whole process can be described formally as follows. 
Let $L=\{l_0,\ldots,l_k\}$  be the set of available object labels. Let $A=\{a_0,\ldots,a_j\}$ be a set of algorithms performing the mapping $M:R^n\rightarrow L^n$ such that $\forall x,y\in I,\;\; p_{x,y}\in L$ with $p_{x,y}$ is a pixel located at coordinates $x$ and $y$. Let $F_i$ be a set of features extracted from input image $I_i$ and let $S(F_i, -, A)$ be an algorithm selection mechanics realizing the mapping $\lambda: F \rightarrow A$. 

The processing starts by extracting features $F_i$ from input image $I_i$ (Figure~\ref{fig:system} box 1) which are used by the algorithm selector $S(F_i,A)$ (Figure~\ref{fig:system} step 2) to determine the most appropriate algorithm $a_j$. The resulting symbolic segmentation is a pixel-wise labeling of the initial image: $L_i\vert \forall x,y\; p_{xy}\in L $. 

\begin{figure}[bht]
	\centering
	\includegraphics[width=0.5\linewidth]{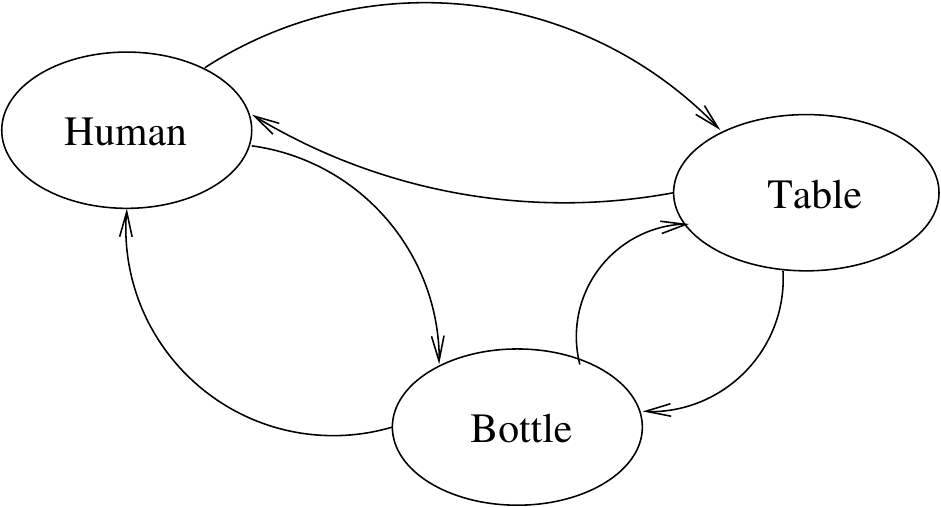}
	\caption{\label{fig:graph1} Example of a multi-relational graph between three objects}
\end{figure}

From $L_i$, a fully connected multi-relational graph (Figure~\ref{fig:graph1}) $ML_{G}$ representing the interaction between the various detected objects is obtained by representing each recognized objects by a node $V_l$ and relations between each objects are represented by an edge $E_{lk}$. 

The relations represented by the edges $E_{lk}$ are obtained from co-occurrence statistics generated from the training data for the following relations: relative position of the center of the gravity, relative size \em rs \em and proximity \em rp\em. The relative size is represented by four coefficients \em l \em (left), \em r \em (right), \em u \em (up) and \em d \em (down). Each of the relation values obtained from the co-occurrence statistics is calculated for each pair of objects (eq.~\ref{eq:edge}).
\begin{equation}
	\{l,r,u,d,rs,rp\} = \{L\circ l_{ij},R\circ r_{ij},U\circ u_{ij},D\circ d_{ij},S\circ rs_{ij},P\circ rp_{ij}\}
	\label{eq:edge}
\end{equation}
The right side of eq.~\ref{eq:edge} shows that values of each relation is obtained by comparing (shown as $\circ$ in eq.~\ref{eq:edge}) the value of a particular ratio calculated from two objects $i$ and $j$ from the semantic segmentation $L_{i}$, with the co-occurrence matrix coefficient $L_{op}$  representing the relation average value. Thus for instance $l=L\circ l_{ij}$ represents the relation that object $i$ is \em left of \em object $j$. 

The vector at each edge of the $ML_G$ has three components: the position $p$ represented by a weighted equation $p=\frac{w_0l+w_1r+w_2u+w_3d}{4}$, size $rs$ and the proximity $rp$. The coefficients $w_0,\ldots,w_3$ are binary and are obtained by simply comparing the centers of gravity of all pairs of detected objects. Example of a multi-relational graph for three objects is shown in Figure~\ref{fig:graph1}. In Figure~\ref{fig:graph1}  the values are given for the outside link between each nodes. Notice that between each two nodes two edges with opposite orientation are created; the graph is anti-symmetric. This means that if for the outside link the value of the position is given by $p_o = 1*l+0*r+1*u+0*d$ then for the inside link it will be $p_i=0*l+1*r+0*u+1*d$ and $p_o=\frac{1}{p_i}$. Similarly for size $rs_o = \frac{1}{rs_i}$. The only value that is the same for both links is the proximity parameter $rp_i = rp_o$. Proximity is calculated as the average value of how many times two objects are in direct contacts or not. 

The high-level description represented by $ML_G$ is analyzed for symbolic contradiction (Figure~\ref{fig:system}  step 5) by looking at the values of each relations. 
By analyzing the coefficients in each of the edges of the $L_i$ the existence of a symbolic contradiction is determined using eq.~\ref{eq:contr}. 
\begin{equation}
C = \begin{cases} \text{True}& \text{ if } \frac{w_0l+w_1r+w_2u+w_3d+rs+rs}{6} < \theta\\  \text{False}&O.W.\end{cases}
	\label{eq:contr}
\end{equation}
The threshold coefficient $\theta$ is determined experimentally so that during the training of the contradiction detection the accuracy on the validation data set is at least 90\%.

The multi-relational graph representation of the existence of a contradiction means that any of the elements in a graph can be modified in order to solve the contradiction. For instance, assume that in Figure~\ref{fig:graph1} a contradiction is detected between Human and Bottle and between Table and Human. 

To determine which of the nodes $V_l$ are to be considered for modification, a contradiction histogram $C_l$ with bins representing each considered vertex $V_l$ and the value of each bin represents the number of times node $V_l$ was present in a relation resulting in a contradiction. The candidate node for replacement by a new hypothesis is then $max_l(C_l)$. If such choice does not result in a single node, nodes with the highest occurrence during the contradiction check are all verified one by one. 

If the contradiction is detected a hypothesis is generated for each contradiction by finding such a node $V_l$ that will maximize all of its edges values $E_{lk}$. This is equivalent to generating a hypothesis generated by the largest co-occurrence statistics given the symbolic segmentation for all but one regions being fixed. Similarly to the contradiction, a set of hypotheses for each node $V_l$ is generated in order of relevance. 
\begin{table}[bht]
\centering
\caption{\label{tab:conthyp} Example of a Contradictions obtained from graph in Figure~\ref{fig:graph1} and the corresponding hypotheses.}
\begin{tabular}{|c|c|c|c|}
	\hline
	Label&Contradiction with&Contradiction&Hypothesis\\
	\hline
		Human&Table& Human Too Small&Car\\
		Human&Bottle& Wrong Position& Bird\\
		Table&Human& Table Too Large& Chair\\
		Bottle&Human& Bottle Wrong Position& Bird\\
	\hline
\end{tabular}
\end{table}

When the multi-relational graph has one of its nodes replaced a new graph is obtained. The new node is the hypothesis $H_l$ that is used as input to the algorithm selector together with features extracted from the region of the contradiction $F_l$: $a_l = S(F_l, H_l, A)$. The new resulting semantic segmentation $L_j$ is merged with the initial segmentation $L_i$ as follows: $L_m = L_c \bowtie L_i$ and with $L_c\vert \forall x,y \in C,\; p_{xy}\in L$ with $\bowtie$ means that the pixel area corresponding to the contradiction in segmentation $L_i$ was replaced by the labeling from segmentation $L_c$ (Figure~\ref{fig:system} step 4). 


Once the new result is obtained a new multi-relational graph is created and another iteration begins. This iterative approach continues until there are no more contradictions or when no more algorithms can be selected.

This platform will be referred to as Automated Selection Method (ASM) as it incrementally changes the high level description of the input image.
\begin{figure}[bht]
	\centering
\fbox{
   \includegraphics[width=0.45\linewidth]{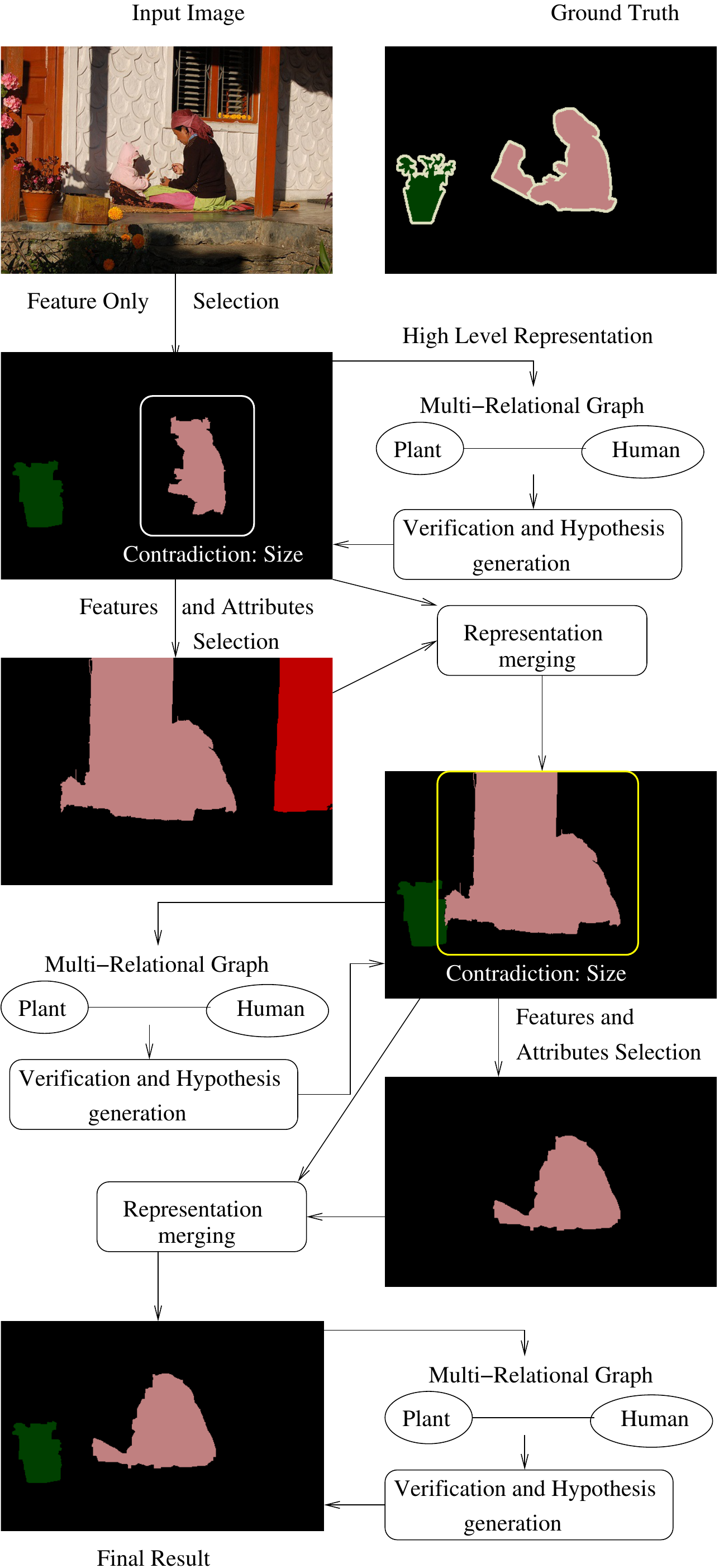}
}
   \caption{\label{fig:exm}Exemplar processing of an input image by the ASM platform}
\end{figure}


An example of ASM processing an image is shown in Figure~\ref{fig:exm}.

\subsection{Hypothesis Representation}

The verification and the hypothesis generation suffers from the almost infinite available hypotheses. This is a problem of representation because a very large number of hypotheses would require a constantly growing amount of space for storage. 

In order to solve this problem the generated hypothesis is represented by a set of attributes allowing to represent all available hypotheses with a constant size of representation. The attributes used are extracted using the \em regionprops \em function in Matlab. Ten out of all attributes have been experimentally determined to allow complete and crisp distinction between the possible labels. Note that the available hypotheses are only from the set of possible labels of the used dataset. 
Each attribute is calculated as an average of the values extracted from all objects encountered in the training data set. The extracted features from the image are together with the attributes clustered and are used for training and testing of the algorithm selection. 



\section{Experiments}
\label{sec:exp}
To evaluate the proposed framework we used the VOC2012 data and up to five different algorithms for symbolic segmentation~\cite{ladicky:10,carreira:12,bharath:14,long:15,mostajabi:14} called ALE, CPMC, SDS, CNET1 and CNET2  respectively in this paper. Each of the algorithms use similar or none preprocessing, different segmentation and similar classification machine learning based object recognition. Initially some tests are performed on three algorithms ALE, CPMC and SDS (referred to as set-3) and the set of all five algorithms will be referred to as set-5.

\subsection{Training of the Automated Selection Method}
\label{sec:selects}
\begin{figure}[bht]
	\centering
	\subfloat[][]{
	\includegraphics[width=0.4\textwidth]{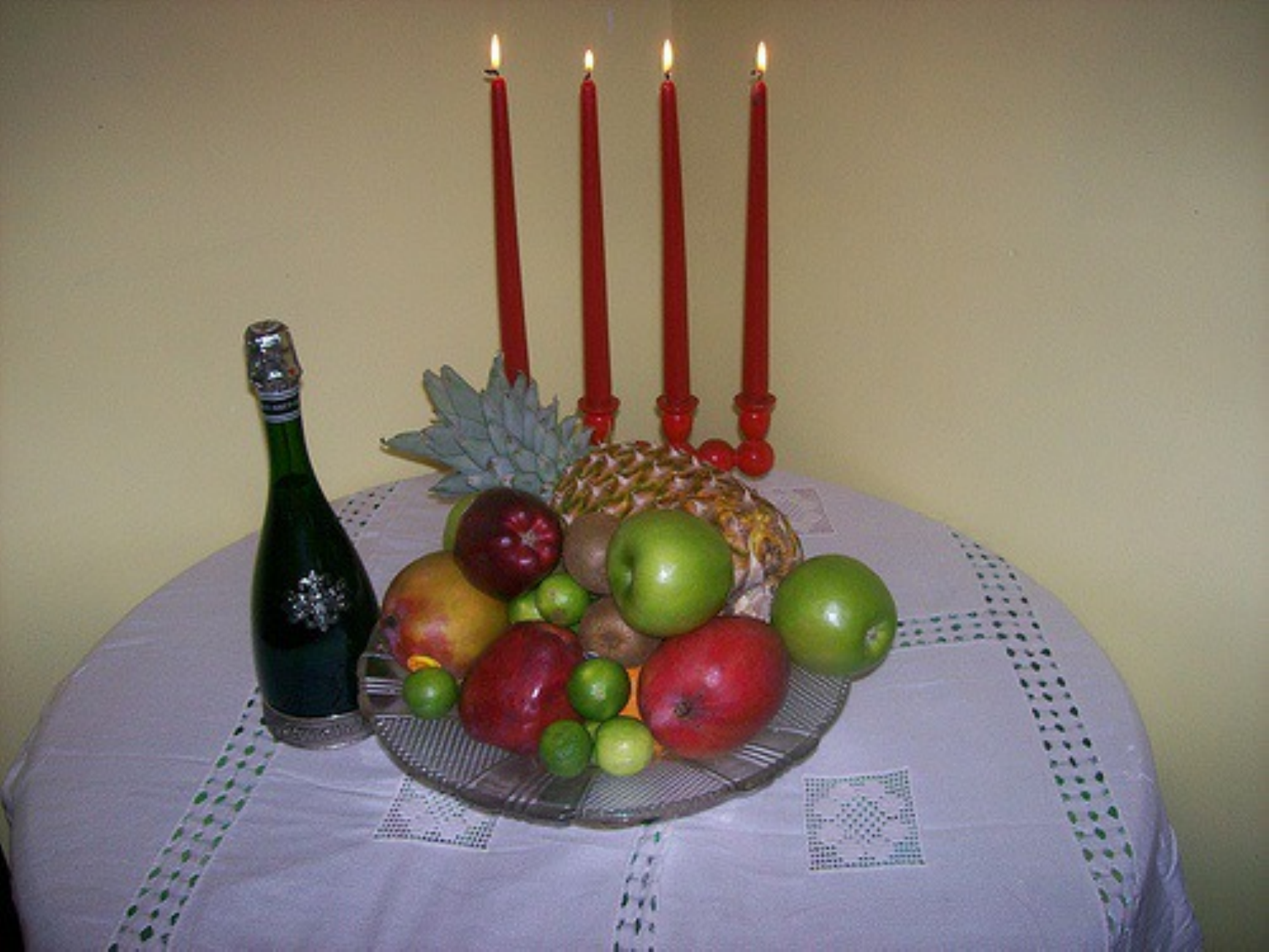}}
	~ 
	\subfloat[][]{
	\includegraphics[width=0.4\textwidth]{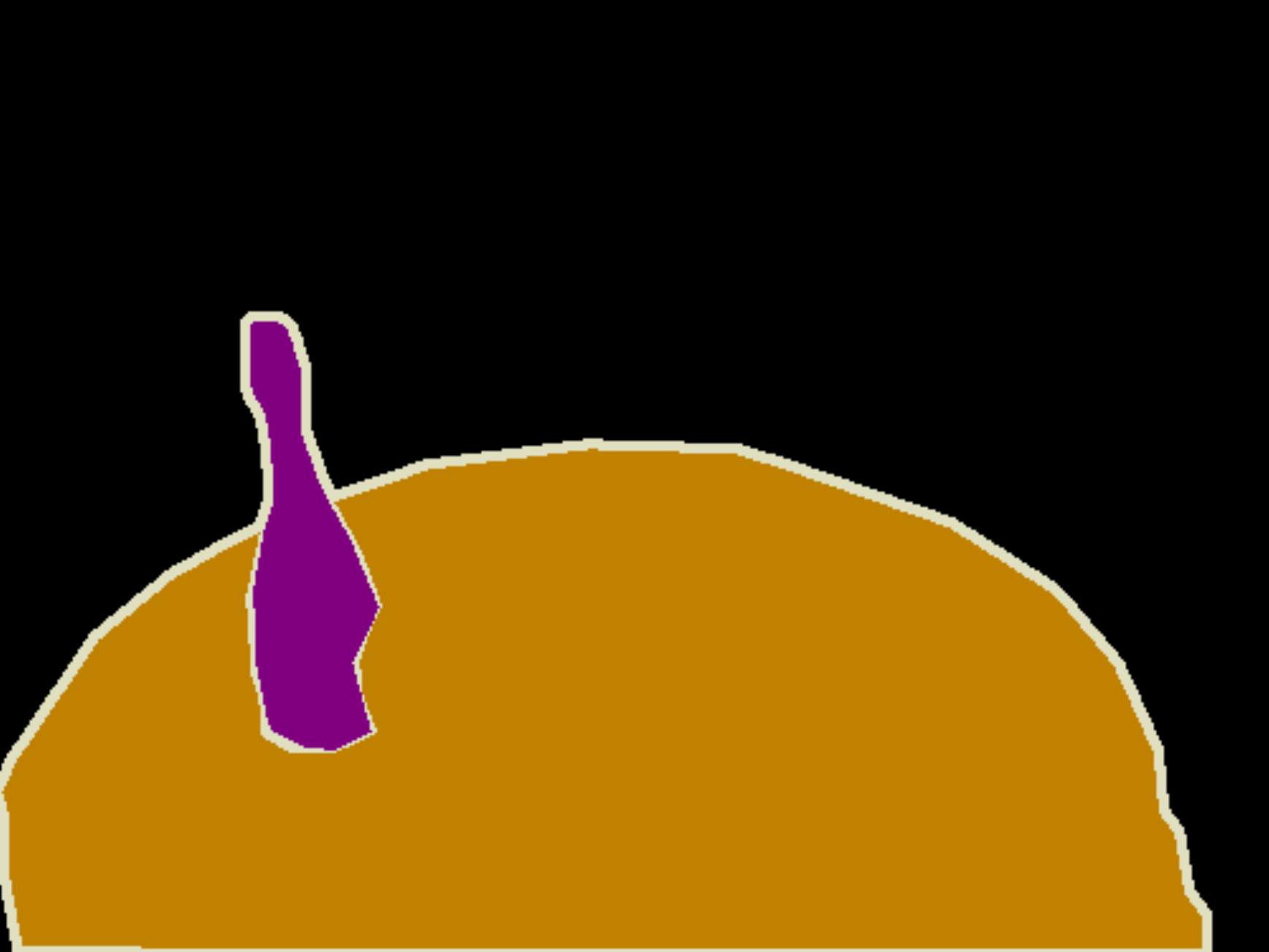}}\\
	\subfloat[][]{
	\includegraphics[width=0.45\textwidth]{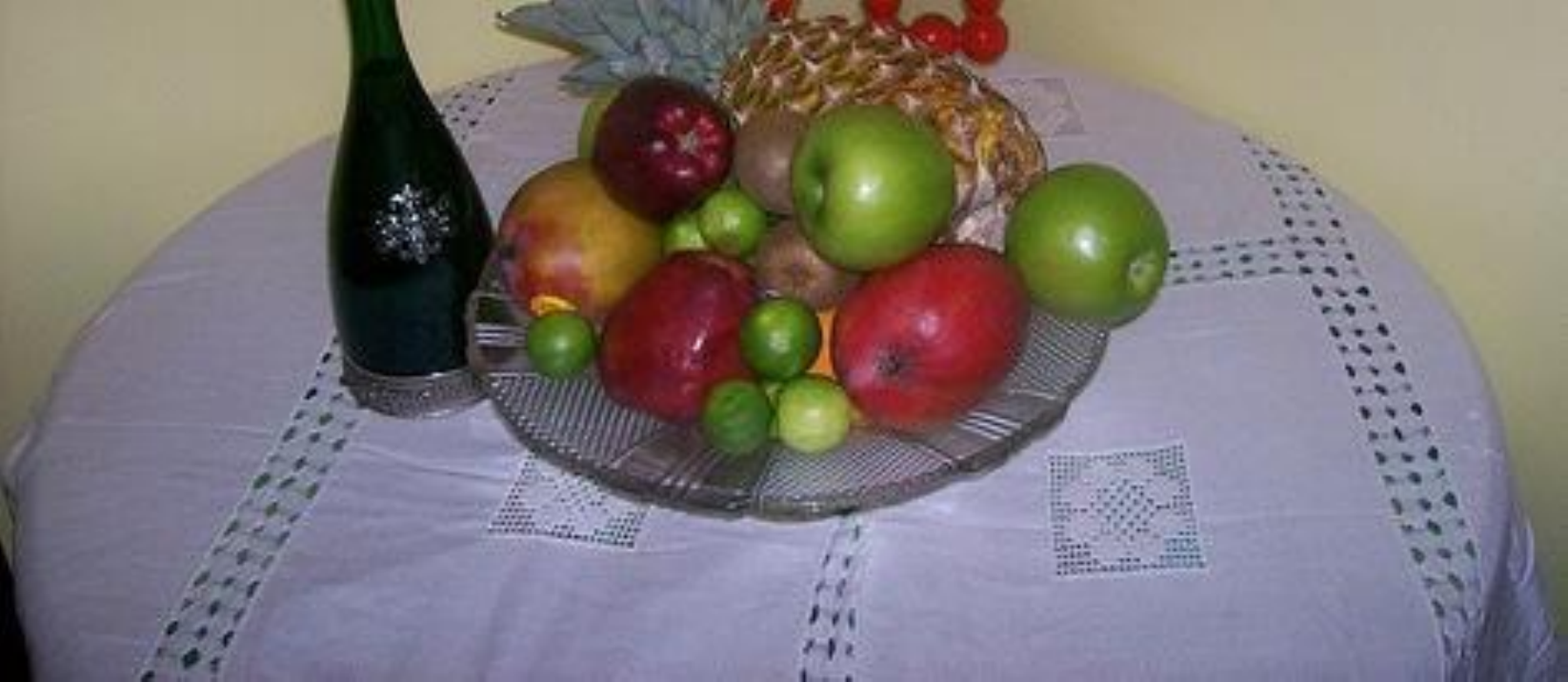}}
	~ 
	\subfloat[][]{
		\centering
	\includegraphics[width=0.1\textwidth]{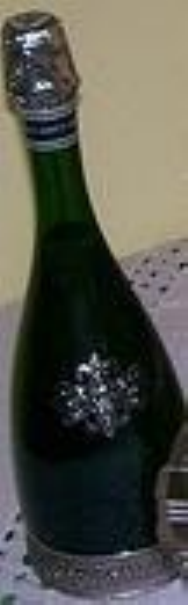}}
	\caption{\label{fig:bboxs} Example of an (a) whole input image, (b) ground truth, (c) bounding box for table and (d) bounding box for bottle.}
\end{figure}

The algorithm selection methods have all been trained with sub-images of bounding boxes containing objects or with whole images. If the given selection method contained two separate algorithm selectors both data sets have been used otherwise only the bounding boxes have been used for training. Example of a whole image, ground truth and the bounding boxes containing objects are shown in Figure~\ref{fig:bboxs}. The reason for using two sets of training data sets is to accommodate the two types of algorithm selection; the first one being $S(F_i,-,A)$ and corresponds to algorithm selection with input features extracted from the whole image. The second algorithm selection is $S(F_i,H_t,A)$ that is used to iteratively improve the initial result of the first selected algorithm.

The features $F_i$ extracted from the image used by the algorithm selector are: color intensity, fft coefficients, brightness intensity, image contrast, and features obtained using a convolutional network implemented in the CAFFE framework~\cite{jia:14}. 

The first training data set $T_f$ is equivalent to the VOC2012 training data set and for each of the images from this set only features are extracted. The feature vector contains all together 7856 feature values composed from histograms of all extracted features.

The second training data set $T_a$ is created from bounding boxes of around the semantic segmentations in the training set of VOC2012 data set. Same features as in $T_f$ are extracted but additionally a set of attributes are used to uniquely describe the correct semantic segmentation.

Three algorithm selection methods have been tested and evaluated: SVM, BN and ANN. 

\subsubsection{SVM:}
SVM  is a very efficient machine learning method however is not well suited for handling unknown information~\cite{pelckmans:05}: in the first iteration of ASM only features are used (at this point in the processing no hypothesis is available yet) while in all subsequent iterations features and hypothesis are used. To solve this problem we experimentally determine that patching approach~\cite{mallison:03} outperformed all other solutions such as the two separate SVMs. Using the patching approach, whenever the hypothesis is not available, (attributes of an image could not be obtained because hypothesis was not generated or it is unknown) the attributes values were generated by the average of the available values. 

\subsubsection{Bayesian Network:}
In the case of the BN only $T_a$ is used for training as the BN is well suited to handle missing input values: once trained with both attributes and features, BN can be used for algorithm selection also using only features. However the BN approach requires deterministic input values - observations. Because most of the features extracted are continuous values within a certain range it is necessary to cluster the data to discrete values. The clusterization is done using an equivalent ranges for each value given by~(\ref{eq:ranges}). 
\begin{equation}
	r_i = [(max_f-min_f)/k*(i-1), (max_f-min_f)/k*(i)]
	\label{eq:ranges}
\end{equation}
with $k$ being the number of values that this value is intended to have and $i$ is the $i-th$ range.
The motivation for using BN is due to the ability of using hierarchy of information and thus to reduce the complexity of learning.

\subsubsection{Artificial Neural Network:}
The last algorithm selection method was a fully connected feed forward neural network. Unlike in the SVM case, the ANN approach was using two ANNs: one trained only using features and the other was trained using features and hypothesis attributes. The ANN used was a single hidden layer multi-layered Perceptron with sigmoid activation function.



\subsubsection{Preference Rules}

In addition to the main three categories of algorithm selectors an additional mechanism that can be used only when the hypothesis is available for the feedback was evaluated. This mechanism is a set of rules that are direct generalization of the results of per-class accuracy of each semantic segmentation algorithm. Simply put, if an algorithm $a_j$ during the training has the highest accuracy in segmenting class $c_i$, then whenever the hypothesis $c_i$ is generated during the testing the algorithm $a_j$ will be used.

\subsection{Testing of the Platform}

The testing of the system was done over a subset of images from the VOC2012 validation data set; only images that contain at least two objects in the ground truth have been used. Using such images both levels of algorithm selection as well as result merging was evaluated. As introduced in Section~\ref{sec:algosel} the high level verification requires multiple objects detections in one image. Images with single objects only cannot be verified at this stage of the ASM platform as the contradiction generation and verification is based on the analysis of inter-objects relations. The co-occurrence matrices for each of the properties defined for each edge of the high-level representation graph were trained on the VOC2012 train data set and the accuracy were verified on $\frac{1}{3}$ of the images from the validation data set.

Consequently the results obtained as baseline accuracy of the used algorithms is different than the results originally reported by the algorithms' respective authors.

At first we evaluate the algorithm selector ability to classify the images according to which algorithm results in best symbolic segmentation. To evaluate the classification power of all three (plus the rules) algorithm selectors we analyzed results  multi-class classification (using the set-3 algorithms). Then the whole system is analyzed by looking at the resulting data. 

\begin{table}[bht]
	\centering
	\caption{\label{tab:algoselacc} Comparison of average accuracy of the different possible algorithm selection methods for the set-3 and set-6 lgorithm set}
	\begin{tabular}{|c|c|c|}
		\hline
		Algorithm Selection Method&Accuracy Set-3&Accuracy Set-6\\
		\hline
		BN&45\%&31\%\\
		SVM&50\%&39\%\\
		SVM-SVM&55\%&45\%\\
		SVM-Rules&60\%&56\%\\
		ANN-SVM&62\%&57\%\\
		ANN-Rules&64\%&60\%\\
		\hline
	\end{tabular}
\end{table}
Table~\ref{tab:algoselacc} shows the results of comparing six different algorithm selection mechanisms based on the four introduced methods in Section~\ref{sec:selects}. Notice that the accuracy is still relatively low at this stage and only some of the best performing combinations of algorithm selection methods are shown. For instance the approach for algorithm selection combining two ANNs is not shown as its results were similar to that of the ANN-SVM selection method. The accuracy of algorithm selection also decreases proportionally to the increasing number of available algorithms.

\begin{figure*}
	\centering
	\includegraphics[width=\textwidth]{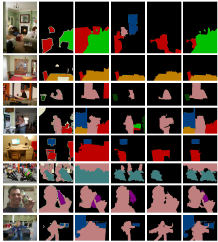}
	\caption{\label{fig:results}Selected results form the ASM platform. Each row represents one particular input. Column (1) shows the input image, column (2) shows the human generated ground truth, columns (3)-(5) shows the results of the three available algorithms in order ~\cite{ladicky:10,bharath:14,carreira:12} and last column shows the result obtained by ASM platform.  }
\end{figure*}

Some examples of processing are shown in Figure~\ref{fig:results}. Notice that despite the low accuracy a number of images are improved by selecting the regions from each algorithm. 

Table~\ref{tab:asmres} compares the results of the ASM platform when used with three and six algorithms for semantic segmentation. In both caes the ASM has the same setting, same input features, same attributes but the test images might vary slightly. The reason for the image variation is because the contradiction and hypothesis generation accuracy is not 100\% and thus for each image the contradiction might or might not be detected. The important fact to be observed is that while with three algorithms available ASM outperformed the best by 2\% while with six availabel algorithms the ASM was better only by 0.15\%. This confirms the fact that with increasing number of algorithms the algorithm selection accuracy must be preserved.  
\begin{table}[bht]
	\centering
	\caption{\label{tab:asmres} Comparison of the ASM method using three and six segementation algorithms.}
	\begin{tabular}{|c|c|c|}
		\hline
		\# of Algorithms&Best Algorithm&ASM\\
		\hline
		3&ALE (38.42)&40.65\\
		6&WDO (69.85)&67\\
		\hline
	\end{tabular}
\end{table}

To see how well the ASM approach is performing we compare the average precision of each category of class. Comparison of each algorithm's results is shown in Table~\ref{tab:evalstats}. As can be seen the ASM framework outperformed the highest classes precision only in six classes of objects: the car, horse, motorbike, sheep, sofa and train. For the rest of the categories the ASM approach was able to outperform most of the algorithms but one. However despite hte relatively low accuracy of the ASM the result is still better on the average. The results in Table~\ref{tab:evalstats} are the results of using the ANN-Rules based approach. Notice that for the classes airplane and boat none of the algorithms detected and correctly segmented these objects in the testing data. This is possible because the test data used (a subset of the validation data from the VOC 2012 data set) can contain exactly those images where none of the algorithms successfully process the images.

Notice that in at least one case the ASM algorithm has equal segmentation accuracy to the best algorithm. This is due to the fact that for this particular class of objects the ASM converged to a single algorithm. Example of this case is the class sheep in Table~\ref{tab:evalstats}.

\begin{table}[bht]
	\centering
	\caption{\label{tab:evalstats} Example of per class accuracies with three algorithms on the test dataset}
	\begin{tabular}{|c|c|c|c|c|c|}
	\hline
	Accuracy for&\multicolumn{4}{|c|}{\multirow{4}{*}{Algorithm Names}}&Best\\
	each class &\multicolumn{4}{|c|}{}&Algorithm\\
	(intersection &\multicolumn{4}{|c|}{}&\\
	/union measure)&\multicolumn{4}{|c|}{}&\\
	\cline{2-5}
	&\texttt{ALE}&\texttt{COMP}&\texttt{CPMC}&\texttt{ASM}&\\

	\hline
	background:     &54.878\%& 80.061\%& 77.478\%& 62.157\%&COMP\\
	airplane:       &--     &--     &--     &--     &--\\
	bicycle:        &26.799\%& 31.913\%& 14.515\%& 27.624\%&COMP\\
	bird:           &22.070\%& 37.042\%& 59.947\%& 21.932\%&CPMC\\
	boat:           &--     &--     &--     &--     &--\\
	bottle:         &37.445\%& 50.990\%& 39.280\%& 50.226\%&COMP\\
	bus:            &44.212\%& 12.034\%& 71.156\%& 45.412\%&CPMC\\
	car:            &52.788\%& 34.924\%& 31.873\%& 56.241\%&ASM\\
	cat:            &63.939\%& 65.552\%& 62.707\%& 63.802\%&COMP\\
	chair:          &19.113\%& 22.355\%&  7.800\%& 19.014\%&COMP\\
	cow:            &33.093\%&    0.0\%&    0.0\%& 30.991\%&ALE\\
	diningtable:    &39.155\%& 50.907\%& 23.997\%& 40.169\%&COMP\\
	dog:            &60.085\%& 49.253\%& 49.827\%& 59.148\%&ALE\\
	horse:          &46.406\%& 27.761\%& 27.155\%& 47.128\%&ASM\\
	motorbike:      &61.154\%& 28.477\%& 33.949\%& 61.697\%&ASM\\
	person:         &46.362\%& 63.940\%& 46.068\%& 57.947\%&COMP\\
	pottedplant:    &25.762\%& 36.391\%& 25.045\%& 23.245\%&COMP\\
	sheep:          &69.008\%& 66.129\%& 27.191\%& 69.008\%&ASM\\
	sofa:           &29.672\%& 17.062\%& 11.806\%& 29.702\%&ASM\\
	train:          &43.602\%&  0.000\%& 28.651\%& 51.174\%&ASM\\
	tvmonitor:      &31.320\%& 62.904\%& 53.201\%& 37.091\%&COMP\\
	\hline
	Average accuracy: &38.422\%& 35.128\%& 32.935\%& 40.653\%&ASM\\
	\hline
	\end{tabular}
\end{table}

\subsubsection{Contradiction and Hypothesis Generation Accuracy}

According to the schematic of the ASM platform the low accuracy of the algorithm selector could be compensated by a stronger verification and reasoning mechanism. Consider the third row in Figure~\ref{fig:results}. A better reasoning procedure would lead to a result as shown in the hypothetical and ideal case shown in Figure~\ref{fig:exm} rather to the result shown in the last column of the third row in Figure~\ref{fig:results}. The simplest heuristics that would prevent replacing regions directly reducing the f-value could increase the overall result without any significant computational overhead. Similar heuristics for improbable regions removal can also be implemented in parallel to the co-occurrence statistics. Thus even a relatively inaccurate algorithm selection with combined with simple high level verification would lead to better results. 

With respect to the general low level of accuracy of the contradiction detection and hypothesis generation, the co-occurrence statistics are only a very simple first step into building a more general model of reasoning on the partially known symbolic content. The reasoning on the labeled 2D shapes is very simple and does not account for more general context from the background, situation describing more complex behavioral interaction and so on. 
\begin{table}[bht]
	\centering
	\caption{\label{tab:hypoth} The accuracy of the co-occurrence statistics based contradiction detection and hypothesis generation}
	\begin{tabular}{|c|c|}
		\hline
		Total Accuracy of Contradiction/Hypothesis&0.610\\
		False positives&407\\
		True negatives&613\\
		Total Samples&1020\\
		\hline
	\end{tabular}
\end{table}


\section{Conclusion}
\label{sec:con}
In this paper we introduced a soft computing approach to the semantic segmentation problem. The method is based on an algorithm selection platform with the target to increase the quality of the result by reasoning on the content of algorithms outputs. The ASM platform for image understanding iteratively improves the high level understanding and even with a very weak algorithm selector can outperform in many cases the best algorithm by combining the best results of each available algorithm. 

In the future several direct extensions and improvements are planned to the ASM platform. First the algorithm selection accuracy must be improved. Second the high level verification also requires a more robust method of contradiction detection and hypothesis generation. Co-occurrence statistics are not sufficient because their dependence on the training data. Finally the result merging requires more flexible and robust mechanism in order to avoid decrease in result quality.
{\small
\bibliographystyle{ieee}
\bibliography{./eccv2016submission}

\begin{thebibliography}{10}\itemsep=-1pt

\bibitem{ali:14}
A.~Ali, M.~Couceiro, A.~Hassanien, M.~Tolba, and V.~Snasel.
\newblock Fuzzy c-means based liver ct image segmentation with optimum number
  of clusters.
\newblock In P.~Kroemer, A.~Abraham, and V.~Snasil, editors, {\em Proceedings
  of the Fifth International Conference on Innovations in Bio-Inspired
  Computing and Applications IBICA 2014}, volume 303 of {\em Advances in
  Intelligent Systems and Computing}, pages 131--139. Springer International
  Publishing, 2014.

\bibitem{ali:11}
R.~Ali, M.~Gooding, T.~Szilagyi, B.~Vojnovic, M.~Christlieb, and M.~Brady.
\newblock Automatic segmentation of adherent biological cell boundaries and
  nuclei from brightfield microscopy images.
\newblock 23(4):607--621, 2011.

\bibitem{arbalez:06}
P.~Arbelaez.
\newblock Boundary extraction in natural images using ultrametric contour maps.
\newblock In {\em Computer Vision and Pattern Recognition Workshop}, pages
  182--190, june 2006.

\bibitem{arbelaez:12}
P.~Arbelaez, B.~Hariharan, C.~Gu, S.~Gupta, L.~Bourdev, and J.~Malik.
\newblock Finding animals: Semantic segmentation using regions and parts.
\newblock In {\em International Conference on Computer Vision and Pattern
  Recognition}, pages 3378 -- 3385, 2012.

\bibitem{carreira:12}
J.~Carreira, F.~Li, and C.~Sminchisescu.
\newblock Object recognition by sequential figure-ground ranking.
\newblock {\em International Journal of Computer Vision}, 98(3):243--262, 2012.

\bibitem{chen:14}
L.~Chen, G.~Papandreou, I.~Kokkinos, K.~Murphy, and A.~Yuille.
\newblock Semantic image segmentation with deep convolutional nets and fully
  connected crfs.
\newblock {\em CoRR}, abs/1412.7062, 2014.

\bibitem{ciresan:12}
U.~Ciresan, D.~andMeier and J.~Schmidhuber.
\newblock Multi-column deep neural networks for image classification.
\newblock Technical report, IDSIA, 2012.

\bibitem{everingham:10}
M.~Everingham, L.~Van~Gool, C.~K.~I. Williams, J.~Winn, and A.~Zisserman.
\newblock The pascal visual object classes (voc) challenge.
\newblock {\em International Journal of Computer Vision}, 88(2):303--338, June
  2010.

\bibitem{felzenswalb:10}
P.~Felzenszwalb, R.~Girshick, D.~McAllester, and D.~Ramanan.
\newblock Object detection with discriminatively trained part based models.
\newblock {\em IEEE Transactions on Pattern Analysis and Machine Intelligence},
  32(9), 2010.

\bibitem{ferryman:05}
J.~Ferryman, M.~Borg, D.~Thirde, F.~Fusier, V.~Valentin, F.~Bremond,
  M.~Thonnat, J.~Aguilera, and M.~Kampel.
\newblock Automated scene understanding for airport aprons.
\newblock In {\em Proceedings of 18th Australian Joint Conference on Artificial
  Intelligence}, pages 593--603, Sidney, Australia, 2005. Springer-Verlag.

\bibitem{gelasca:08}
E.~Gelasca, J.~Byun, B.~Obara, and B.~Manjunath.
\newblock Evaluation and benchmark for biological image segmentation.
\newblock In {\em Proceedings of the International conference on Image
  Processing}, 2008.

\bibitem{bharath:14}
B.~Hariharan, P.~Arbel\'{a}ez, R.~Girshick, and J.~Malik.
\newblock Simultaneous detection and segmentation.
\newblock In {\em European Conference on Computer Vision}, pages 297--312,
  2014.

\bibitem{heikkila:04}
J.~Heikkila and O.~Silven.
\newblock A real-time system for monitoring of cyclists and pedestrians.
\newblock {\em Image and Vision Computing}, 22(7):563 -- 570, 2004.
\newblock Visual Surveillance.

\bibitem{hoiem:08}
D.~Hoeim, A.~A. Efros, and M.~Hebert.
\newblock Closing the loop on scene interpretation.
\newblock In {\em Proc. Computer Vision and Pattern Recognition (CVPR)}, pages
  1 -- 8, June 2008.

\bibitem{jia:14}
Y.~Jia, E.~Shelhamer, J.~Donahue, S.~Karayev, J.~Long, R.~Girshick,
  S.~Guadarrama, and T.~Darrell.
\newblock Caffe: Convolutional architecture for fast feature embedding.
\newblock {\em arXiv preprint arXiv:1408.5093}, 2014.

\bibitem{jung:06}
H.~Jung, D.~Kim, P.~Yoon, and J.~Kim.
\newblock Structure analysis based parking slot marking recognition for
  semi-automatic parking system.
\newblock In D.-Y. Yeung, J.~Kwok, A.~Fred, F.~Roli, and D.~de~Ridder, editors,
  {\em Structural, Syntactic, and Statistical Pattern Recognition}, volume 4109
  of {\em Lecture Notes in Computer Science}, pages 384--393. Springer Berlin
  Heidelberg, 2006.

\bibitem{kolmogorov:07}
V.~Kolmogorov, Y.~Boykov, and C.~Rother.
\newblock Applications of parametric maxflow in computer vision.
\newblock In {\em In Proceedings of the International Conference on Computer
  Vision}, 2007.

\bibitem{ladicky:10}
L.~Ladicky, C.~Russell, P.~Kohli, and P.~Torr.
\newblock Graph cut based inference with co-occurrence statistics.
\newblock In {\em Proceedings of the 11th European conference on Computer
  vision}, pages 239--253, 2010.

\bibitem{lathen:10}
G.~Lathen.
\newblock Segmentation methods for digital image analysis: Blood vessels,
  multi-scale filtering, and level set methods, 2010.

\bibitem{leibe:08}
B.~Leibe, A.~Leonardis, and B.~Schiele.
\newblock Robust object detection with interleaved categorization and
  segmentation.
\newblock {\em International Journal of Computer Vision}, 77:259­289, 2008.

\bibitem{li:09}
L.-J. Li, R.~Socher, and L.~Fei-Fei.
\newblock Towards total scene understanding:classification, annotation and
  segmentation in an automatic framework.
\newblock In {\em Computer Vision and Pattern Recognition (CVPR)}, pages 2036
  -- 2043, 2009.

\bibitem{long:15}
J.~Long, E.~Shelhamer, and T.~Darrell.
\newblock Fully convolutional networks for semantic segmentation.
\newblock In {\em CVPR}, 2015.

\bibitem{lowe:99}
D.~G. Lowe.
\newblock Object recognition from local scale-invariant features.
\newblock In {\em Proceedings of the International Conference on Computer
  Vision}, 1999.

\bibitem{lukac:11d}
M.~Lukac and M.~Kameyama.
\newblock Adaptive functional module selectiopn using machine learning:
  Framework for intelligent robotics.
\newblock In {\em Proceedings of the SICE}, 2011.

\bibitem{lukac:13}
M.~Lukac, M.~Kameyama, and K.~Hiura.
\newblock Natural image understanding using algorithm selection and high level
  feedback.
\newblock In {\em SPIE Intelligent Robots and Computer Vision XXX: algorithms
  and Techniques}, volume 8662, page 86620D, 2013.

\bibitem{lukac:12b}
M.~Lukac, R.~Tanizawa, and M.~Kameyama.
\newblock Machine learning based adaptive contour detection using algorithm
  selection and image splitting.
\newblock {\em Interdisciplinary Information Sciences}, 18(2):123--134, 2012.

\bibitem{maire:08}
M.~Maire, P.~Arbelaez, C.~Fowlkes, and J.~Malik.
\newblock Using contours to detect and localize junctions in natural images.
\newblock In {\em Computer Vision and Pattern Recognition}, pages 1--8, 2008.

\bibitem{malik:99}
J.~Malik, S.~Belongie, J.~Shi, and T.~Leung.
\newblock Textons, contours and regions: Cue combination in image segmentation.
\newblock In {\em International Conference on Computer Vision}, 1999.

\bibitem{mallison:03}
H.~Mallinson and A.~Gammerman.
\newblock Imputation using support vector machines, 2003.

\bibitem{martin:01}
M.~Martin, C.~Fowlkes, D.~Tal, and J.~Malik.
\newblock A database of human segmented natural images and its application to
  evaluating segmentation algorithms and measuring ecological statistics.
\newblock In {\em International Conference on Computer Vision}, volume~2, pages
  416 -- 423, July 2001.

\bibitem{meijering:12}
E.~Meijering.
\newblock Cell segmentation: 50 years down the road.
\newblock 29(5):140--145, 2012.

\bibitem{mharib:12}
A.~Mharib, A.~Ramli, S.~Mashohor, and R.~Mahmood.
\newblock Survey on liver ct image segmentation methods.
\newblock 37(2):83--95, 2012.

\bibitem{mostajabi:14}
M.~Mostajabi, P.~Yadollahpour, and G.~Shakhnarovich.
\newblock Feedforward semantic segmentation with zoom-out features.
\newblock {\em CoRR}, abs/1412.0774, 2014.

\bibitem{oliva:01}
A.~Oliva and A.~Torralba.
\newblock Modeling the shape of the scene: a holistic representation of the
  spatial envelope.
\newblock {\em International Journal of Computer Vision}, 42(3):145--175, 2001.

\bibitem{pelckmans:05}
K.~Pelckmans, J.~De~Brabanter, J.~Suykens, and B.~De~Moor.
\newblock Handling missing values in support vector machine classifiers.
\newblock 18(5-6):684--692, 2005.

\bibitem{peng:08}
B.~Peng and V.~Veksler.
\newblock Parameter selection for graph cut based image segmentation.
\newblock In {\em British Conference on Computer Vision}, pages 16.1--16.10,
  2008.

\bibitem{perera:06}
A.~G.~A. Perera, G.~Brooksby, A.~Hoogs, and G.~Doretto.
\newblock Moving object segmentation using scene understanding.
\newblock In {\em Conference on Computer Vision and Pattern Recognition}, 2006.

\bibitem{price:10}
B.~Price, B.~Morse, and S.~Cohen.
\newblock Geodesic graph cut for interactive image segmentation.
\newblock In {\em In Proceedings of the International Conference on Computer
  Vision and Pattern Recognition}, 2010.

\bibitem{rice:76}
J.~Rice.
\newblock The algorithm selection problem.
\newblock {\em Advances in Computers}, 15:65­118, 1976.

\bibitem{sharma:10}
N.~Sharma and L.~M. Aggarwal.
\newblock Automated medical image segmentation techniques.
\newblock 1(35):3–14, 2010.

\bibitem{shi:00}
J.~Shi and J.~Malik.
\newblock Normalized cuts and image segmentation.
\newblock {\em IEEE Transactions on Pattern Analysis and Machine Intelligence},
  22(8):888--905, 2000.

\bibitem{takemoto:09}
S.~Takemoto and H.~Yokota.
\newblock Algorithm selection for intracellular image segmentation based on
  region similarity.
\newblock In {\em Ninth International Conference on Intelligent Systems Design
  and Applications}, pages 1413 -- 1418, 2009.

\bibitem{tu:02}
Z.~Tu and S.-C. Zhu.
\newblock Image segmentation by data-driven markov chain monte carlo.
\newblock {\em Pattern Analysis and Machine Intelligence, IEEE Transactions
  on}, 24(5):657--673, May 2002.

\bibitem{yong:03}
X.~Yong, D.~Feng, and Z.~Rongchun.
\newblock Optimal selection of image segmentation algorithms based on
  performance prediction.
\newblock In {\em Proceedings of the Pan-Sydney Area Workshop on Visual
  Information Processing (VIP2003)}, pages 105--108, 2003.

\bibitem{zavitz:14}
E.~Zavitz and L.~J. Baker.
\newblock Higher order of image structure enables boundary segmentation in the
  absence of luminance or contrast cues.
\newblock 4(14):1--14, 2014.

\end{thebibliography}
}

\end{document}